\documentclass[letterpaper]{article}

\usepackage[authoryear,round]{natbib}
\usepackage{alifeconf}  
\usepackage{url}
\usepackage[colorlinks=true,allcolors=blue]{hyperref}
\usepackage{booktabs}
\usepackage{amsmath} 
\usepackage{amssymb}
\usepackage{lipsum}
\usepackage{dblfloatfix}
\usepackage{cuted}
\usepackage{caption}
\usepackage{xcolor}
\usepackage{algorithm}
\usepackage{algpseudocode}
\usepackage{svg}
\usepackage{graphicx}

\newcommand\blfootnote[1]{%
  \begingroup
  \renewcommand\thefootnote{}\footnote{#1}%
  \addtocounter{footnote}{-1}%
  \endgroup
}

\title{
The Artificial Experimentalist: Discovery and Control of Self-Organizing Phenomena with Autotelic Reinforcement Learning

}

\author{
    Marko Cvjetko$^{1}$,
    Benedikt Hartl$^{2}$,
    Michael Levin$^{2, 3}$,
    Clément Moulin-Frier$^{4}$,
    Pierre-Yves Oudeyer$^{1}$ \\
    \mbox{}\\
    $^1$Inria Centre at the University of Bordeaux, Bordeaux, France \\
    $^2$Allen Discovery Center at Tufts University, Medford, MA, USA \\
    $^3$Wyss Institute for Biologically Inspired Engineering at Harvard University, Boston, MA, USA\\
    $^4$Inria, INSA Lyon, CITI, UR3720, 69621 Villeurbanne, France \\
    Corresponding Author: \href{mailto:marko.cvjetko@inria.fr}{marko.cvjetko@inria.fr}
} 

\begin{document}

\maketitle

\begin{abstract}
Existing methods for exploring cellular automata and other complex systems mostly operate in open loop: they set initial conditions, execute a full simulation, and observe the outcome, without intervening during execution. We introduce a closed-loop framework based on autotelic reinforcement learning, in which an agent autonomously samples diverse goals and learns a goal-conditioned policy to intervene in a complex system through minimal, local perturbations. We instantiate this framework on Lenia, a continuous cellular automaton known for life-like self-organizing patterns, in an agentic system we call CARL, and demonstrate three capabilities. First, CARL discovers stable solitons across a wide range of Lenia update rules at a higher rate than heuristic baselines. Second, it learns to steer the movement direction of existing solitons with few interventions, showing that CARL can control self-organizing patterns, not only create them. Third, humans can use the agents to guide solitons through maze environments in real time by specifying high-level directional commands that the agents translate into low-level interventions. Trained across diverse goals, update rules, and random initial states, the agents acquire policies that generalize zero-shot to various out-of-distribution conditions. These results suggest a path toward artificial experimentalist agents that, autonomously or with human guidance, discover and control emergent phenomena in complex systems.

\end{abstract}


{\raggedright \textbf{Companion website and code available at:} \url{https://developmentalsystems.org/carl/}\par}

\blfootnote{\textcopyright  2026 [Marko Cvjetko, Benedikt Hartl, Michael Levin, Clément Moulin-Frier, Pierre-Yves Oudeyer]. Published under a Creative Commons Attribution 4.0 International (CC BY 4.0) license.}

\section{Introduction}

One of the central endeavors of science is understanding complex systems at all scales of organization, from elementary physics to astronomy, from molecular biology to ecology. Two general goals drive this research: (1) explaining and discovering the diverse phenomena that emerge in complex systems, and (2) learning to control complex systems toward desired states, ideally with minimal effort. 

In biomedicine, for instance, the goal is not continuous intervention, but the restoration of healthy, self-sustaining dynamics. Rather than controlling individual components, we aim to guide systems back into stable regimes in which they can maintain their function autonomously.
However, identifying such interventions is inherently challenging, as system behavior arises from interactions across many scales. This raises a fundamental question: how can we systematically control systems whose internal dynamics are complex or only partially understood \citep{Levin2023DarwinsAgentialMaterial, levin20254MultiscaleBody}?


Computational approaches are essential for pursuing these goals, as they enable us to simulate complex systems \textit{in silico}. Cellular automata (CAs) have long served as standard models for studying self-organization, and recent continuous extensions such as Lenia~\citep{chanLeniaExpandedUniverse2020a} produce increasingly complex and life-like patterns, making them ideal testbeds for discovering and controlling emergent phenomena. Because the space of behaviors a CA can produce is vast and difficult to anticipate, researchers have developed a range of methods to explore it (see Related Work). Most of these, however, operate in open loop: parameters and initial conditions are chosen upfront, with no interaction during rollout.

This stands in contrast to how humans typically engage with complex systems: continually observing and interacting with the system in real time to form an intuition of its causal dynamics (e.g.,\ a gardener continuously pruning, watering, and reshaping a garden as it grows). As systems grow in complexity, however, effective intervention becomes increasingly difficult. Nonlinear interactions, feedback loops, and delayed effects make human intuition prone to systematic biases, and modern challenges — particularly in biomedical, ecological, or economic contexts — quickly reach the limit of human modeling capabilities~\citep{Tversky1974}.

\begin{figure}[t]
    \begin{tabular}{cc}

        \includegraphics[width=1.0\columnwidth]{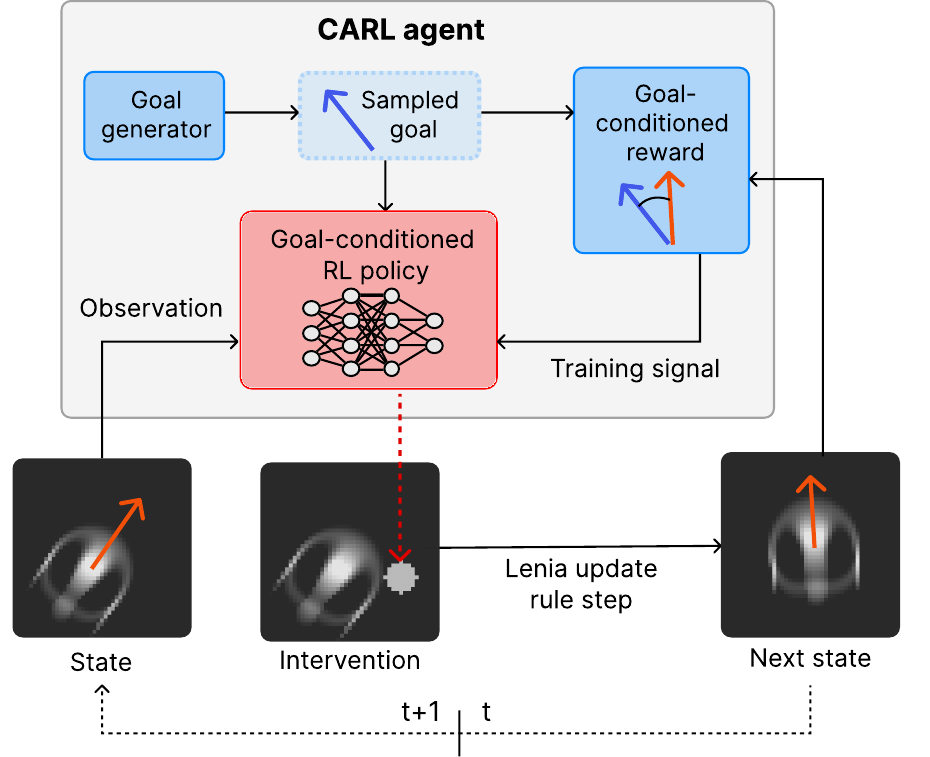} &

    \end{tabular}
    \caption{CARL: an autotelic RL agent that learns to drive the evolution of Lenia toward desired dynamics. At the start of an episode, CARL generates a goal, e.g., a desired moving direction of a Lenia pattern. Then, at each step of the episode, CARL perceives the current Lenia state and outputs an action (adding or removing cell activations) conditioned on the goal,
after which the Lenia update rule is applied.}
    \label{fig:CARL}
\end{figure}

To address these gaps, we propose an autotelic reinforcement learning (RL) framework as an interaction-driven approach to translate a diversity of desired experimental outcomes into actionable, causally effective interventions in complex systems. By \textit{autotelic}, we mean an RL agent that autonomously self-generates and learns to achieve diverse goals — the operational counterpart of these outcomes — in the considered complex system \citep{colasAutotelicAgentsIntrinsically2022a}. As a concrete instantiation, we introduce \textbf{CARL} (for \emph{controlling \textbf{C}ellular \textbf{A}utomata with \textbf{R}einforcement \textbf{L}earning}), which learns to intervene on CAs in a closed-loop fashion, continuously observing and shaping their dynamics over time (Fig.~\ref{fig:CARL}).

Through a series of experiments using Lenia as a testbed, we demonstrate that CARL can both discover interesting self-organizing phenomena and control their behavior, with limited interventions and in a sample-efficient manner. Moreover, CARL generalizes well to out-of-distribution scenarios, such as unseen world dynamics and action spaces. Lastly, we show that trained CARL agents can be deployed interactively, enabling humans to specify and adjust high-level goals in real time while the agents translate them into low-level interventions for controlling complex systems.

While the framework is demonstrated on Lenia, it is designed to be system-agnostic. Looking ahead, we hope to extend it to increasingly biologically grounded models. 

\section{Related Work}
\label{sec:related_work}

\subsection{AI for Scientific Discovery}
 
AI is playing a growing role in scientific discovery. It has already driven breakthroughs in domains such as protein structure prediction and combinatorial optimization~\citep{jumperHighlyAccurateProtein2021, fawziDiscoveringFasterMatrix2022}, and recent work aims to go further by automating the scientific method end-to-end, from hypotheses to publication~\citep{luEndtoendAutomationAI2026, zenilFutureFundamentalScience2026}. A key question in these efforts is how to design agents that can autonomously decide what to investigate and how. The autotelic AI paradigm~\citep{colasAutotelicAgentsIntrinsically2022a}, in which agents set their own goals and learn to achieve them, offers a natural framework for this — capturing the core loop of scientific inquiry: formulating questions and developing the skills to answer them. Population-based autotelic methods have been successfully applied in several scientific contexts, including the automated discovery of protocell behaviors~\citep{grizouCuriousFormulationRobot2020} and the exploration of gene regulatory networks~\citep{etcheverryAIdrivenAutomatedDiscovery2025}.

\subsection{Cellular Automata (CAs)}
 
CAs are dynamical systems consisting of grids of cells whose states are updated based on local neighborhoods. Despite their simplicity, CAs can produce remarkably complex phenomena, making them useful both as models of real-world processes (in ecology, urban development, physics) and as objects of study in their own right. In artificial life, foundational contributions include the work of \citet{turing1952}, \citet{vonNeumann1966}, \citet{barricelli1963numericalI, barricelli1963numericalII}, \citet{langton1986}, and \citet{wolfram1983CAs}, who studied morphogenesis, self-replication, evolution, and complexity.  CAs have recently seen a revival with prominent continuous models such as Lenia and Neural Cellular Automata~\citep{chanLeniaBiologyArtificial2019, chanLeniaExpandedUniverse2020a, mordvintsevGrowingNeuralCellular2020}, and extensions incorporating mass conservation that promote evolutionary phenomena~\citep{plantecFlowLeniaEmergentEvolutionary2025, papadopoulosMaCEGeneralMass2025}. These more expressive CAs have garnered broad interest since they give rise to a plethora of self-organizing patterns that appear increasingly life-like, resembling artificial organisms and ecosystems~\citep{hartlNeuralCellularAutomata2025}.


\subsection{Automated Exploration of Cellular Automata}

Over the years, many methods have been used to illuminate the range of possible CA behaviors, including random search, manual tuning, and hand-crafted heuristics. More recently, gradient-based methods have been used to optimize for specific phenomena~\citep{mordvintsevGrowingNeuralCellular2020, miottiDifferentiableLogicCellular2025, hamonDiscoveringSensorimotorAgency2025}, while diversity-driven algorithms --- including novelty search, quality-diversity, and intrinsically motivated goal exploration processes (IMGEPs) --- aim to discover a variety of distinct behaviors ~\citep{reinkeIntrinsicallyMotivatedDiscovery2020a, etcheverryHierarchicallyOrganizedLatent2020a, faldorArtificialOpenEndedEvolution2024, khajehabdollahiExpeditionExpansionLeveraging2025, michelExploringFlowLeniaUniverses2025}. All of these methods, however, operate in open loop, with no ability to intervene during execution.

Some works have moved beyond this limitation. \citet{rainwaterSelfOrganizationPhaseTransitions} studies how external forces can impact the dynamics of Game of Life. \citet{kumarAutomatingSearchArtificial2025} optimize CA rules at specific checkpoints during evolution, though these are planned in advance rather than chosen reactively. \citet{sanchez-fiblaCooperativeControlEnvironmental2024} train agents in a CA reproducing forest-fire dynamics to manage resource acquisition with environmental extremes. \citet{earleAutoverseEvolvableGame2024} train embodied agents in evolvable CA-based game environments.

\section{Method}
 
\subsection{General Framework}
 
We formalize a framework based on autotelic reinforcement learning \citep{colasAutotelicAgentsIntrinsically2022a} for discovering and controlling phenomena in complex systems. In the first phase, an autotelic agent is trained to achieve a diversity of goals; in the second, the learned goal-conditioned policy serves as a high-level interface to produce self-organizing patterns and control them in real time. The framework is designed to be general, abstracting away the specifics of any particular system. Instantiating it requires defining three components: a \emph{complex system} whose dynamics are to be studied, an \emph{intervention space} through which a goal-conditioned RL policy can intervene in the system, and a \emph{task specification} that defines the goals the policy must learn to achieve.
 
\paragraph{Complex system.} We define a complex system as a tuple $(\mathcal{S}, F)$, where $\mathcal{S}$ is a state space and $F: \mathcal{S} \rightarrow \mathcal{S}$ is an update rule that governs the system's dynamics. At each discrete time step, the system evolves as $\mathbf{s}^{t+1} = F(\mathbf{s}^t)$. We make no assumptions about $F$ beyond the ability to simulate it; it may be deterministic or stochastic, continuous or discrete, and may operate over spatial grids, graphs, particle systems, or other structures.
 
\paragraph{Intervention space.} An intervention is a modification to the system state. We define an intervention function $\alpha: \mathcal{S} \times \mathcal{A} \rightarrow \mathcal{S}$, where $\mathcal{A}$ is the set of available actions. Each action $a \in \mathcal{A}$ produces a perturbation to the current state. The action space can optionally include a \emph{no-op} action, leaving the system unmodified. Crucially, interventions are intended to be small relative to the system --- the agent nudges the system rather than rewriting the whole state.

\paragraph{Task specification.} A task is defined by a goal space $\mathcal{G}$ and a goal-conditioned reward function $r: \mathcal{S}^{\leq T} \times \mathcal{G} \rightarrow \mathbb{R}$, where $\mathcal{S}^{\leq T}$ denotes sequences of states of length $\leq T$. At the start of each training episode, a goal $g \sim p(\mathcal{G})$ is sampled from a predefined distribution. The reward $r(\mathbf{s}^{0:t}, g)$ measures the degree to which the behavior of the system aligns with the goal. Importantly, goals are not limited to a single state: they can include properties of the trajectory, the system's update rules, and constraints on intervention effort (e.g.,\ adding action costs). By conditioning on goals sampled from this rich space, a policy must generalize across diverse objectives.

\paragraph{The loop.} Given a complex system $(\mathcal{S}, F)$, an intervention function $\alpha$, and a task specification $(\mathcal{G}, r)$, we can train a goal-conditioned policy 
$\pi: \mathcal{S}^{\leq T} \times \mathcal{G}\rightarrow\mathcal{A}$ 
through episodic reinforcement learning, to maximize cumulative reward for any goal $g\in\mathcal{G}$ (see Algorithm \ref{alg:carl}).
 
\begin{algorithm}[H]
\caption{The Autotelic Reinforcement Learning Loop}\label{alg:carl}
\begin{algorithmic}[1]
\For{$n = 0, \ldots, n\_episodes-1$}
    \State Sample goal $g \sim p(\mathcal{G})$ and initial state $\mathbf{s}^0$
    \For{$t = 0, \ldots, T-1$}
        \State Observe $\mathbf{s}^{t-\Delta t:t}$ and $g$ over a time interval $\Delta t$
        \State Select $a^t \sim \pi(\cdot \mid \mathbf{s}^{t-\Delta t:t}, g)$
        \State Apply intervention: $\tilde{\mathbf{s}}^t = \alpha(\mathbf{s}^t, a^t)$
        \State Evolve system for $N$ steps: $\mathbf{s}^{t+1} = F^N(\tilde{\mathbf{s}}^t)$
        \State Receive reward $r(\mathbf{s}^{0:t+1}, g)$
    \EndFor
    \State \textit{(Optional)} Roll out $\mathbf{s}^{T+m} = F(\mathbf{s}^{T+m-1})$ for $m = 1, \ldots, M$ to assess resulting phenomena
\EndFor
\end{algorithmic}
\end{algorithm}

During inference, the goal can change dynamically $g\mapsto g^t$, which enables a human user to control the complex system in real time by issuing high-level commands that the trained policy translates into low-level interventions.

\subsection{Instantiation: CARL}

We instantiate CARL (Fig. \ref{fig:CARL}) on Lenia, a continuous generalization of Conway's Game of Life \citep{chanLeniaBiologyArtificial2019}. Lenia's simple update rules produce diverse self-organizing phenomena, making it an ideal testbed for our framework.
 
\paragraph{Lenia $(\mathcal{S}, F)$.} The state is a grid $\mathbf{X}^t \in [0,1]^{H \times W}$ with periodic boundary conditions. The update rule is:
\begin{equation}
    \mathbf{X}^{t+dt} = \left[\mathbf{X}^t + dt\, \varphi(\mathbf{K} * \mathbf{X}^t)\right]_0^1
    \label{eq:lenia_update}
\end{equation}
where $\mathbf{K}$ is a convolutional kernel, $\varphi$ is an element-wise growth function, and $dt$ is the step size.
 
The kernel is defined over a disk of radius $\rho$, partitioned into $b = |\boldsymbol{\beta}|$ concentric rings of equal width with peak values $\boldsymbol{\beta} = (\beta_1, \ldots, \beta_b)$. For a point at normalized distance $r$, the ring index is $i = \min(\lfloor b \cdot r \rfloor, b-1)$ and the local coordinate is $r' = (b \cdot r) \bmod 1$. The unnormalized kernel is:
\begin{equation}
    \tilde{K}(r) = \beta_i \cdot \left(4\, r'(1-r')\right)^4
    \label{eq:kernel}
\end{equation}
and the final kernel is normalized: $\mathbf{K} = \tilde{K} / \sum \tilde{K}$. The growth function maps the convolution output to $[-1, 1]$ via a Gaussian bump:
\begin{equation}
    \varphi(u) = 2\exp\!\left(-\frac{(u - \mu)^2}{2\sigma^2}\right) - 1
    \label{eq:growth}
\end{equation}

Key phenomena of interest in Lenia are \textbf{\emph{solitons}}: localized patterns that persist and often move across the grid. We design our experiments with the intent of showing that CARL can discover new solitons and control their behavior across a range of action costs, across a diversity of update rules, and from procedurally generated initial states. To detect solitons, we apply a simple \textit{soliton filter} inspired by prior work ~\citep{hamonDiscoveringSensorimotorAgency2025, faldorArtificialOpenEndedEvolution2024}: after applying the Lenia update rule for 5,000 steps without intervention, we classify the resulting state as containing a soliton if its total mass is non-zero and is below $10\%$ of the grid capacity. We omit additional checks used in prior work — such as a velocity threshold, mass variability, and robustness testing — since our goal is to evaluate CARL for its ability to create local, persistent patterns of any kind. We visually inspected many simulations that pass the filter and found no false positives (i.e., global, Turing-like patterns were never accepted)\footnote{Many examples of states that pass the soliton filter are available on the companion website}. 
 
\paragraph{Interventions over Lenia $(\mathcal{A}, \alpha)$.} An action is a tuple $a^t = (x^t, y^t, \delta^t)$, where $(x^t, y^t)$ are spatial coordinates and $\delta^t \in \{-1, 0, +1\}$ indicates whether to remove mass, take no action, or add mass. 
An action
modifies all cell values within a radius $R_a$ around $(x^t, y^t)$ by $(\delta^t \cdot M_a)$, with fixed hyperparameters $R_a$ and $M_a$. Values are clipped to $[0,1]$. In all experiments, we use $R_a = 5$ and $M_a = 0.3$ unless stated otherwise.

\subsection{Policy Architecture and Training}

We train policies using Double Deep Q-Networks~\citep{hasseltDeepReinforcementLearning2016}. The network architecture is a U-Net~\citep{ronnebergerUNetConvolutionalNetworks2015}, a fully convolutional network that produces dense Q-value maps over the grid for each action type (add, remove, no-op). This architecture mirrors the structure of CAs: it treats each cell locally while maintaining receptive fields large enough to perceive most or all of the grid. Similar architectures have been used in robotics \citep{zengLearningSynergiesPushing2018, wuSpatialActionMaps2020a}.
 
Observations consist of the last four Lenia grid states. Additional context --- including the goal, action cost coefficient, current episode time step, and Lenia update rule parameters --- is concatenated and provided through FiLM conditioning layers~\citep{perezFiLMVisualReasoning2018}. Both trained agents (introduced below) have $\sim$800K  parameters.
 
We chose an off-policy algorithm for its sample efficiency. Full details of the hyperparameters and network architecture are provided in the code repository.

\section{Experiments}
We demonstrate CARL through three sets of experiments. First, we show that it can create stable solitons across a wide range of Lenia update rules and from procedurally generated initial states, and that trained agents generalize zero-shot to unseen goals, update rules and modified action spaces. Second, we train another agent to steer the movement direction of existing solitons, demonstrating that CARL can not only create self-organizing patterns, but also control them. Third, we show a proof of concept that humans can interact with complex systems through CARL agents by modifying their goals in real time, by having users navigate solitons through a maze using the movement direction agent.

We invite the reader to follow experimental results on the companion website, which contains many video examples.

\subsection{Soliton Creation Task}
 
Rather than searching for solitons directly — which would require defining what constitutes a soliton within the reward signal — we train CARL on a simpler proxy task: maintaining a target mass on the Lenia grid, for a given update rule and action cost. When actions are costly, the agent faces a choice between constantly intervening to hold the mass at the target, and finding a self-sustaining configuration that matches it. Action costs tip the balance toward the latter, making soliton creation an emergent byproduct of reward maximization rather than an explicit objective.

 \begin{figure*}[t]
    \centering
    \includegraphics[width=0.24\textwidth]{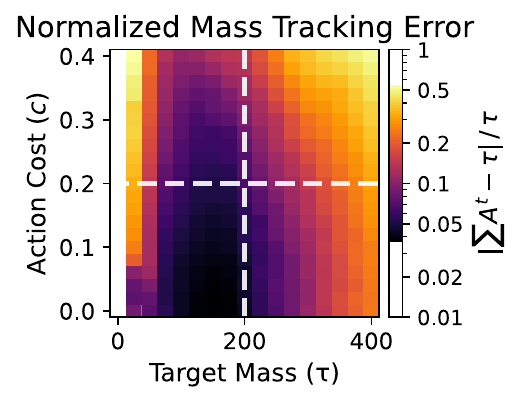}\hfill
    \includegraphics[width=0.24\textwidth]{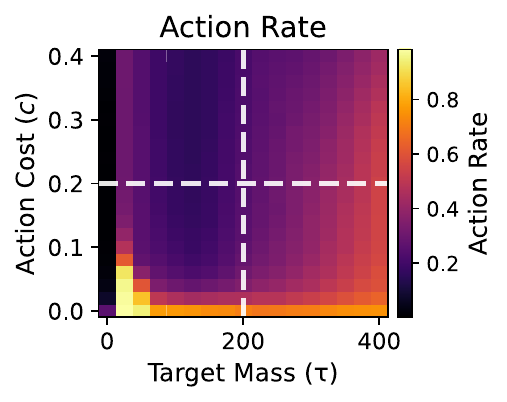}\hfill
    \includegraphics[width=0.24\textwidth]{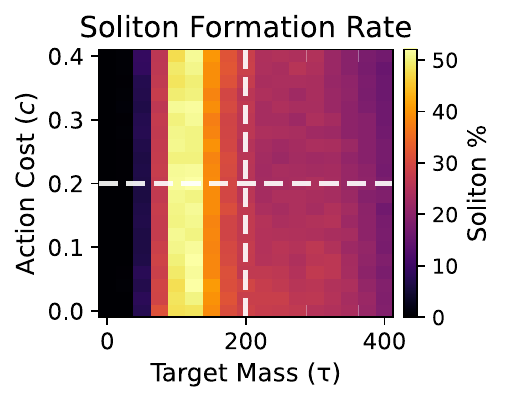}\hfill
    \includegraphics[width=0.24\textwidth]{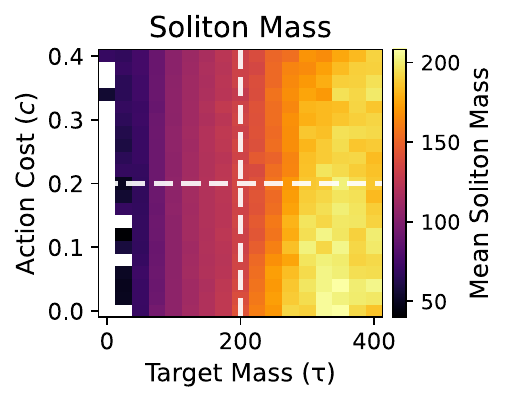}
    \caption{\textbf{Agent behavior and soliton creation metrics} across different
    action costs $c$ and mass targets $\tau$, averaged over all training update
    rules. In every panel, the bottom-left quadrant represents the $\tau$ and $c$
    values seen during training. From left to right: averaged per-step mass error;
    action rate (fraction of steps where the agent intervenes); percentage of
    episodes resulting in solitons; and averaged mass of the generated solitons after
    the soliton-filter period.}
    \label{fig:heatmaps}
\end{figure*}

\paragraph{Goal space and reward.}
The goal space is defined as $\mathcal{G} = \mathcal{T} \times \mathcal{C} \times \Omega$, where $\mathcal{T}$ is the set of target masses, $\mathcal{C}$ the set of action costs, and $\Omega$ the set of update rules. At the start of each episode, CARL samples a goal $g = (\tau, c, \omega)$ uniformly, with $\tau \in [0, 200]$, $c \in [0, 0.2]$, and $\omega$ drawn from a set of training update rules. Sampling the action cost per episode rather than fixing it serves two purposes: it produces a single budget-adaptive policy that can operate across a spectrum of intervention regimes at deployment, and we speculate it also acts as an exploration mechanism during training — low-cost episodes allow the agent to freely discover viable configurations, while high-cost episodes pressure it to find self-sustaining ones. The reward at each step is:
\begin{equation}
    r = -\sqrt{\frac{|M^t - \tau|}{N}} - c \cdot \mathbf{1}[\delta^t \neq 0],
\end{equation}
where $M^t = \sum_{i,j} X^t_{i,j}$ is the total mass of the grid at time $t$, $N$ is the total number of grid cells, and $\delta^t \in \{-1, 0, +1\}$ is the action type selected by the agent. The first term penalizes deviation from the target mass, while the second penalizes non-trivial interventions, weighted by the sampled action cost $c$. A single policy must therefore learn to act across diverse combinations of target masses, action costs, update rules, and initial conditions.
 
\paragraph{Experimental setup.}
Initial states are procedurally generated by randomly applying 20 actions to an empty grid without rolling out the CA in between, producing diverse unstructured configurations. Each episode step consists of an agent's action followed by a single Lenia update step ($N_{\text{steps}}=1$). Episodes last 150 steps on a $64 \times 64$ grid. The model is trained on $2 \times 10^6$ episode-step transitions. The training set of update rules consists of 85 hand-selected rules supporting diverse solitons discovered by \citet{hudcovaVisualizingStructureLenia2025}.


See our companion website for a
detailed description of 
hyperparameters and included update rules.
Unless stated otherwise, 
evaluation conditions are
run for 16 episodes.

\subsubsection{Mass tracking evaluation.}

 We evaluate the agent on the Cartesian product of target masses
$\tau \in \{0, 25, 50, \ldots, 400\}$, action costs
$c \in \{0, 0.05, 0.1, \ldots, 0.4\}$, and all 85 training update
rules $\omega \in \Omega_{\text{train}}$. For each triplet $(\tau, c, \omega)$,
we run 16 episodes, yielding a three-dimensional evaluation grid.
Both $\tau$ and $c$ extend to twice their training range.
Fig.~\ref{fig:heatmaps} shows projections onto the
$(\tau, c)$ plane, with metrics averaged over update rules.
 

 
The agent tracks the target mass reliably across most of the evaluation range (Fig.~\ref{fig:heatmaps}, left). Performance degrades at boundary values of $\tau$, particularly when action costs are high. This is expected: at both extremes, the system likely lacks stable self-sustaining configurations — mass dissipates at low $\tau$ and grows unboundedly at high $\tau$ — forcing the agent into costly continuous intervention. The action cost also shapes the behavior as intended: when $c > 0$, the agent acts less frequently and relies more on the intrinsic dynamics of the system (Fig.~\ref{fig:heatmaps}, center-left).
 
\subsubsection{Soliton creation.}
 
We now turn to the central question: does the agent produce solitons? To test this, we take the final Lenia grid state from each evaluation episode above, roll it out for 5,000 Lenia steps without any agent intervention, and apply the soliton filter.
 
We observe that the agent produces solitons at a high rate, especially for target masses between $\tau=100$ and $150$ (Fig.~\ref{fig:heatmaps}, center-right), and that the mass of created solitons correlates well with the target (Fig.~\ref{fig:heatmaps}, right).  For very low target masses, almost no episodes yield solitons, consistent with the observation above that such masses cannot persist without constant intervention. For high targets, the agent discovers an interesting strategy: creating several independent solitons whose combined mass matches
the target. Action costs have little impact on the soliton formation rate, which decreases only slightly at very low values. However, we speculate that varying the action cost per episode was nonetheless crucial during training.
 
 
\paragraph{Comparison with baselines.}
We compare CARL against several heuristic baselines across all training update rules, with a fixed action cost of $c = 0.1$ and mass targets $\tau \in \{50, 100, 150, 200\}$. The baselines include: \textbf{No-op} (always selects no-op), \textbf{Random} (random action type and location), \textbf{Mass-based} (adds/removes mass toward target, placed randomly or at locations of existing mass), and \textbf{Mass-based with deadzone} (same, but only acts when mass deviates by more than 10\% from target).
 
\begin{figure}[t]
    \centering
    \includegraphics[width=\columnwidth]{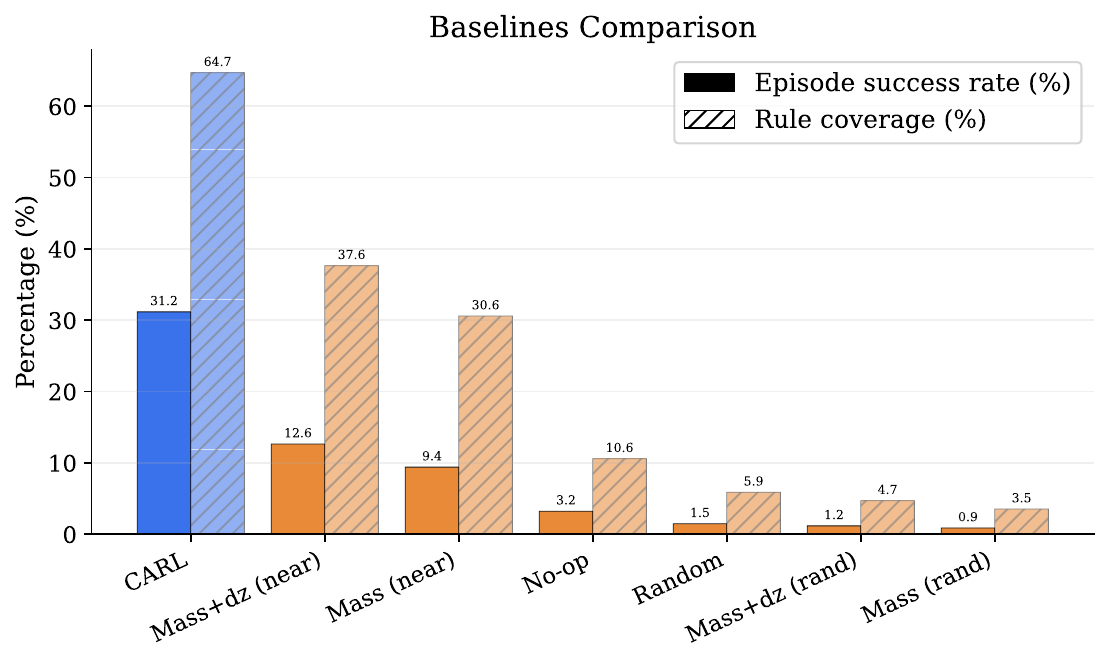}
    \caption{
        CARL compared to baseline methods for soliton creation, averaged across all training update rules with action cost $c = 0.1$ and mass targets $\tau \in \{50, 100, 150, 200\}$.
    }
    \label{fig:baselines}
\end{figure}

CARL outperforms all baselines in both the overall soliton creation rate and in the number of update rules for which at least one soliton is generated (Fig.~\ref{fig:baselines}). The gap is particularly notable against the mass-based heuristics, which have access to the same mass information as CARL but lack spatial awareness: they cannot learn where to place mass to seed a viable pattern. The no-op baseline confirms that solitons rarely arise from random initial conditions alone, underscoring that the agent's interventions are essential.

\subsubsection{Generalization}
 
The results above show that CARL reliably creates solitons under training conditions. We now assess how robust this capability is by testing three axes of generalization: modified action parameters, rescaled update rule kernels, and entirely novel update rules.
 
\begin{figure}[t]
    \centering
    \includegraphics[width=0.53\columnwidth]{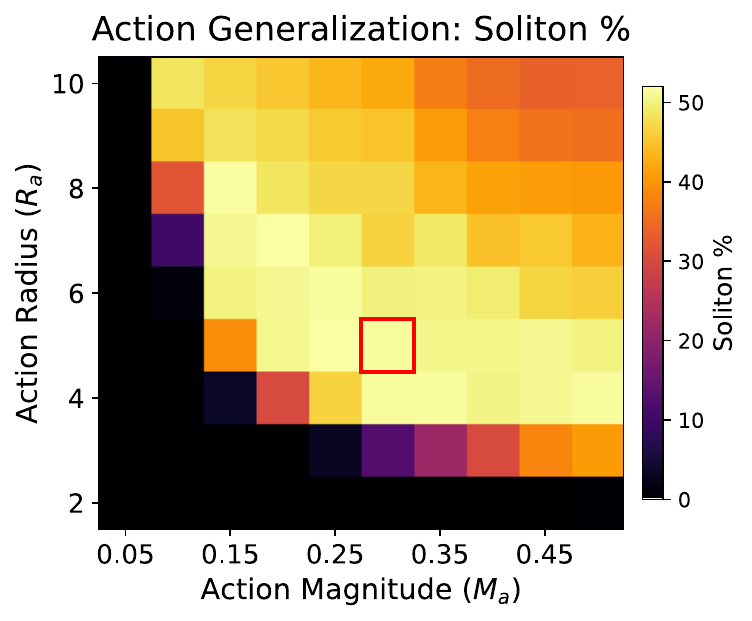}%
    \hfill
    \includegraphics[width=0.44\columnwidth]{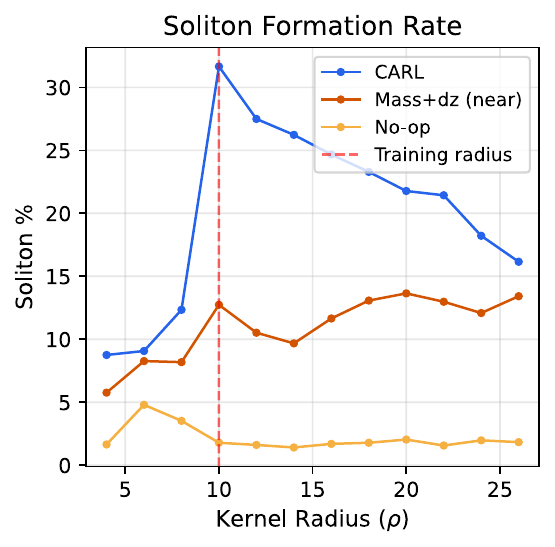}
    \caption{
        Soliton creation rate under modified parameters.
        Left: Varying action radius and magnitude; the red square marks training conditions.
        Right: Soliton creation rate for varying kernel radius $\rho$ contrasting CARL with baselines.
    }
    \label{fig:action_kernel_generalisation}
\end{figure}

\paragraph{Modified action parameters.}
We evaluate the agent when the action hyperparameters --- $R_a$ and $M_a$ --- are changed at test time. Importantly, the agent does not observe these hyperparameters. We fix $\tau = 125$ and $c = 0.1$, conditions that produce solitons reliably under training settings. The agent adapts well to combinations where $R_a \cdot M_a \approx R_a^{\text{train}} \cdot M_a^{\text{train}}$, i.e.,\ when the integrated effect of each action matches that of the training conditions (Fig.~\ref{fig:action_kernel_generalisation}, left), and degrades gracefully away from this curve. Performance drops to zero only for low-impact action hyperparameters, where individual actions are too weak to seed or sustain any mass on the grid.

\paragraph{Rescaled kernel radius.}
We evaluate whether the agent can create solitons when the update rule kernel radius $\rho$ is rescaled. This is a particularly challenging form of generalization: rescaling the kernel does not simply scale the emerging patterns, but can fundamentally alter their behavior due to discretization effects. The same update rule at different kernel radii can produce solitons with different shapes, sizes, movement patterns, and levels of robustness.
 
We deploy the agent across kernel radii $\rho \in \{4, 6, 8, \ldots, 26\}$, all training update rules, target masses $\tau \in \{50, 100, 150, 200\}$, and action costs $c \in \{0.1\}$. When rescaling $\rho$, we proportionally adjust the action radius and grid size (linearly) and the target masses (quadratically), keeping the ratio between action scale, target mass, and pattern size roughly constant. This isolates the effect of rescaled dynamics on agent performance. All rescaled settings produce out-of-distribution values. Since the policy networks are fully convolutional, they can be deployed on different grid sizes without modification.
 
The agent adapts well to scaled Lenia worlds, particularly for up-scaled kernels (Fig.~\ref{fig:action_kernel_generalisation}, right). Although performance drops compared to the training radius, the agent still creates solitons at a relatively high rate, even for kernels double or half the training size. Across all tested radii, the agent's soliton creation rate remains well above the no-op baseline and comparable to or above the best heuristic baseline evaluated at the training radius. The success rate drops more sharply for down-scaled kernels, likely due to discretization effects.

\begin{figure*}[t]
    \centering
    \includegraphics[width=0.42\textwidth]{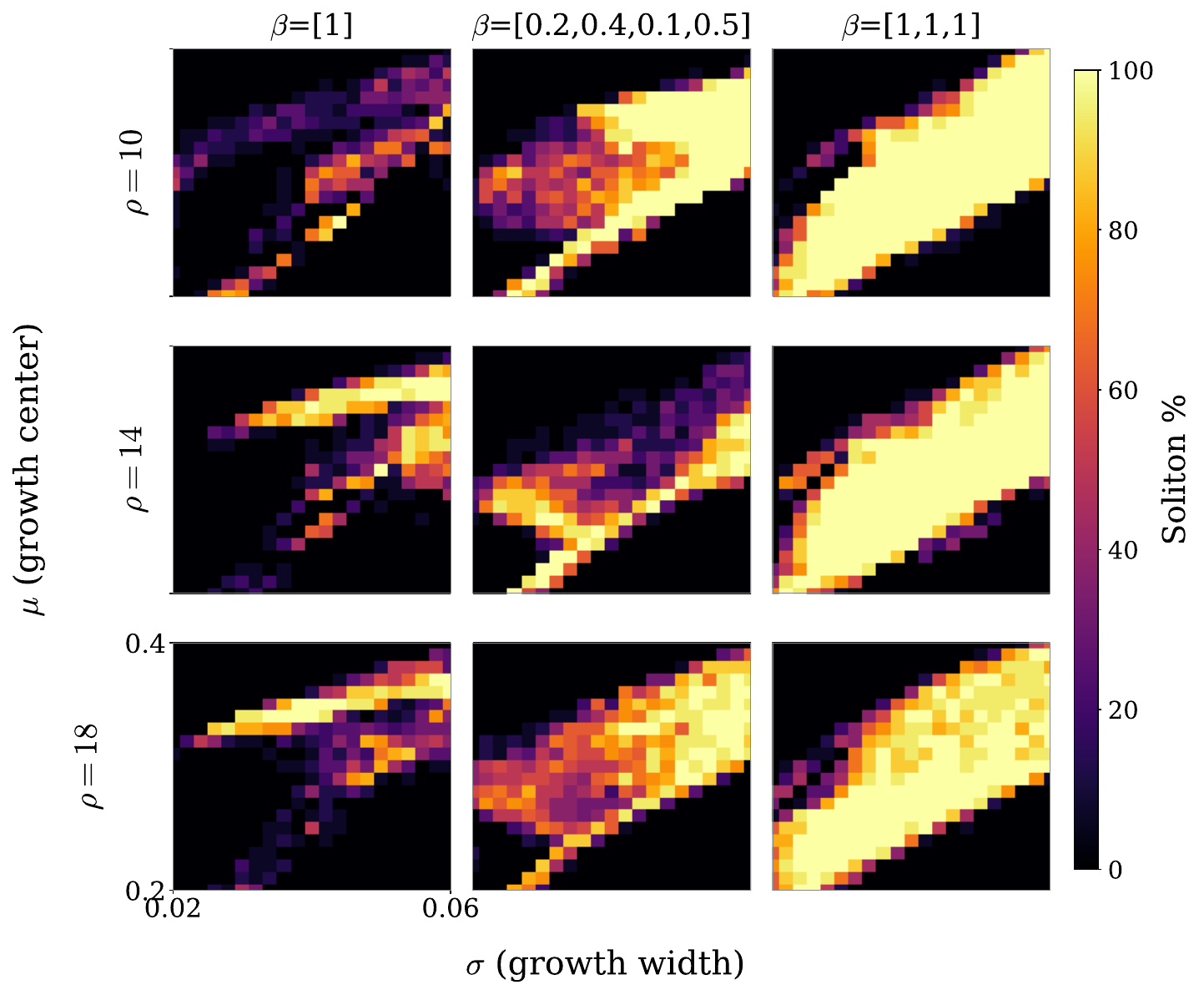}
    \hspace{2em}
    \includegraphics[width=0.42\textwidth]{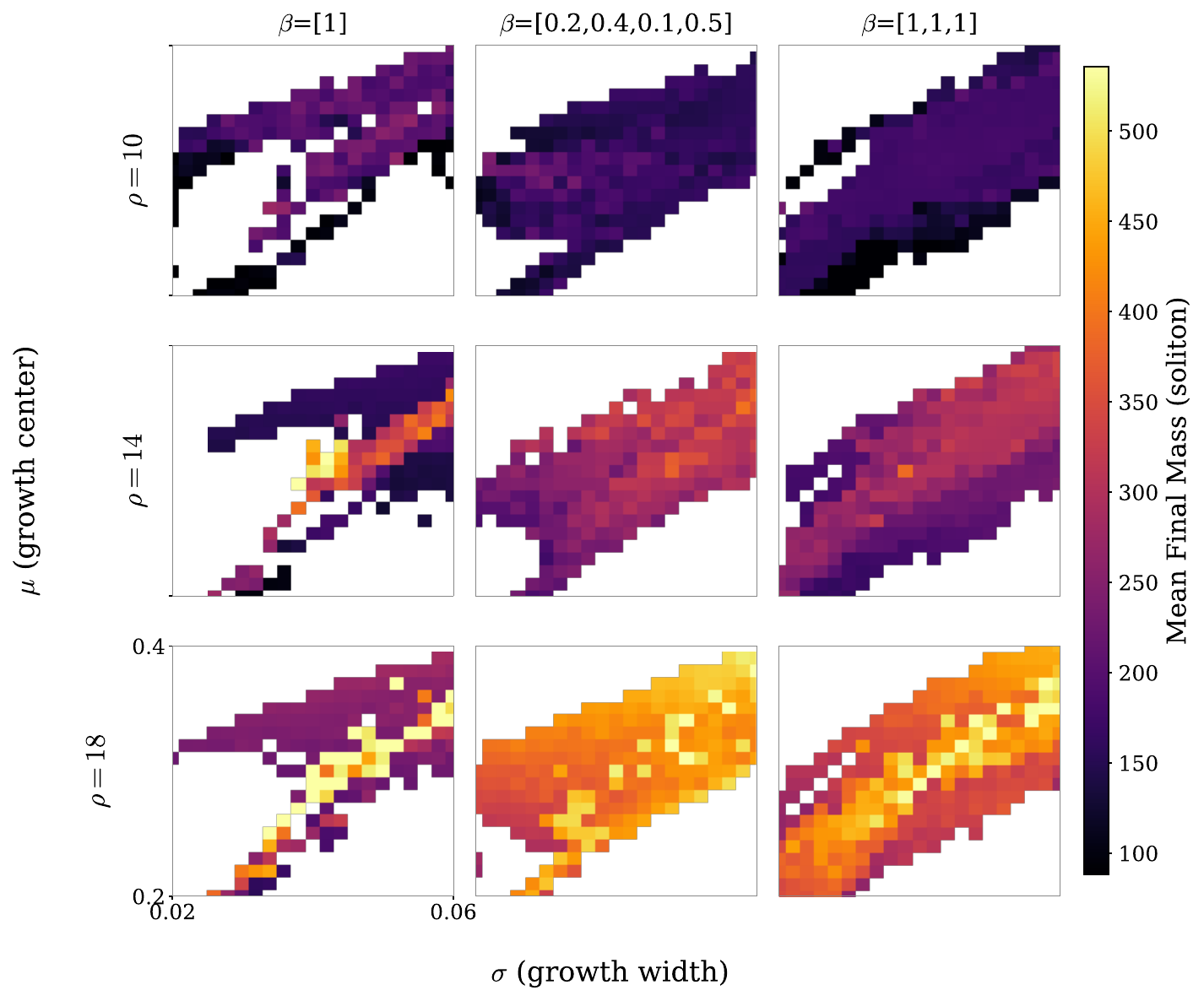}

    \vspace{0.5em}

    \includegraphics[width=0.74\textwidth]{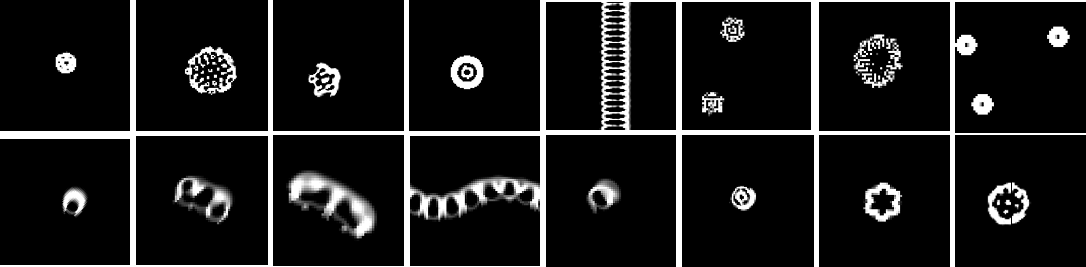}

    \caption{Soliton formation rate maps for novel update rules. Left: likelihood of soliton creation as a function of growth function parameters $(\mu, \sigma)$, for a fixed kernel $\beta$ and radius $\rho$. Right: mean soliton mass under the same conditions. Bottom: examples of solitons discovered in novel update rules.}
    \label{fig:novel_rules_grid_soliton_pct}
\end{figure*}

\paragraph{Novel update rules.}
Finally, we deploy CARL on unseen convolutional kernels. We select \textit{seven} convolutional kernels $\mathbf{K}$ not included in the training set, and sweep across growth function parameters $\mu \in \{0.2, 0.205, \ldots, 0.4\}$ and $\sigma \in \{0.02, 0.022, \ldots, 0.06\}$ with $\tau \in \{50, 100, 150, 200, 250\}$ and $c = 0.1$. 
We conduct these grid searches for kernel radii $\rho \in \{10, 14, 18\}$, rescaling relevant hyperparameters as was done in the previous evaluation. Fig.~\ref{fig:novel_rules_grid_soliton_pct} visualizes 
the resulting space and snapshots of some discovered solitons.
 
The results show that the trained agent can efficiently map novel update rule spaces, identifying which regions of $(\mu, \sigma)$ space support soliton formation. The agent discovers solitons across a range of unseen kernels, though the success rate varies considerably depending on the kernel --- some kernels admit large regions of soliton-supporting parameters, while others are more restrictive. This suggests that a trained CARL agent can serve as a practical tool for rapid exploration of new Lenia update rules for downstream tasks.
 
\subsection{Soliton Direction Task}

To demonstrate that CARL can control self-organizing phenomena, not only create them, we train a new agent on a task where it must steer a soliton toward a target direction. Each episode lasts $200$ steps and begins with a uniformly sampled target direction, an action cost (as before), and a soliton drawn from a set of $48$ that the first experiment's agent discovered 
in 
the kernel-scaling generalization 
test ($\rho = 18$); we use the larger radius because solitons at smaller radii tend to move chaotically. To initialize the state, the sampled soliton is placed on the grid under its corresponding update rule and randomly rotated. A soliton's movement direction is measured by the center-of-mass displacement vector over the last four timesteps. The reward combines two terms: (1) the cosine similarity between current and target direction, and (2) a mass penalty, as in the previous task, with the target mass set to that of the initial state. We train the agent for
$10^6$
episode-step transitions.

We evaluate generalization along two axes: \emph{solitons} and \emph{directions}. The 48 solitons are split into 24 training and 24 holdout (stratified across $(\mu, \sigma)$), and the unit circle of target directions is partitioned into four quadrants, of which only two opposite ones are used during training. This yields a $2 \times 2$ generalization grid with $128$ episodes per cell.

\begin{figure}[t]
    \centering
    \includegraphics[width=0.7\columnwidth]{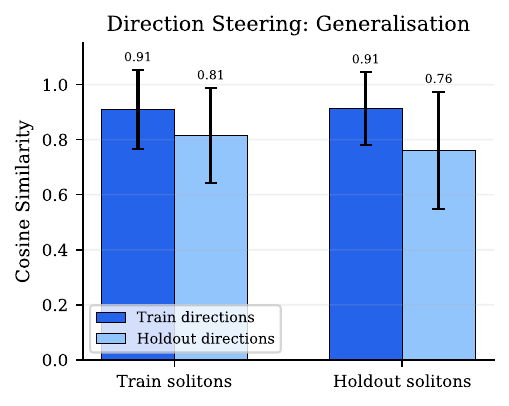}
    \caption{Generalization of the direction-steering agent. Mean cosine similarity between the soliton's movement and the target direction, evaluated on a $2 \times 2$ grid of \{train, holdout\} solitons $\times$ \{train, holdout\} directions. Error bars show one STD across 128 episodes per cell.}
    \label{fig:velocity_generalisation}
\end{figure}

Fig.~\ref{fig:velocity_generalisation} shows the mean cosine similarity between the soliton's center-of-mass displacement and the target direction, averaged over all $200$ episode steps across the four conditions: On training solitons and training directions, the agent achieves a mean cosine similarity of $0.91 \pm 0.14$, indicating strong directional alignment. Performance degrades gracefully under generalization: to $0.81 \pm 0.14$ for unseen directions (training solitons), $0.91 \pm 0.10$ for unseen solitons (training directions), and $0.76 \pm 0.15$ when both are held out. Notably, generalization to unseen solitons within training directions is nearly lossless, suggesting that the steering policy captures direction-dependent strategies that transfer across soliton morphologies. The larger drop for holdout directions indicates that the mapping from direction to intervention pattern is only partially learned.

\subsection{Human-in-the-Loop} 
\begin{figure}[t]
    \centering
    \includegraphics[width=0.58\columnwidth]{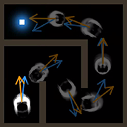}
    \caption{A timelapse of a human guiding a soliton through a maze by modifying the goals of the soliton steering agent in real time. The blue and orange arrows represent target and current movement directions, respectively. Note that the soliton's shape is preserved well throughout the episode.}
    \label{fig:maze_timelapse}
\end{figure}

We demonstrate how trained CARL agents can serve as real-time interfaces for human control. We extend the direction environment with procedurally generated mazes, where walls are regions in which cell values are fixed to zero. A soliton is placed in the maze and the user can modify the agent's goal (target direction) and action cost in real time, steering the soliton through the maze (Fig. \ref{fig:maze_timelapse}).

The agent has no explicit representation of the maze — it perceives only the single-channel Lenia grid, identical to its training setting. Furthermore, the agent was never trained with changing goals, yet it successfully redirects solitons multiple times within a single episode while preserving their coherent shape. Reducing the action cost makes the agent intervene more frequently and advance faster.

Although the agent performs well, several failure modes emerge. Wall collisions can cause the soliton to disintegrate or explode, though some solitons are robust to contact. High action costs can also lead to failure, as interventions become too sparse to maintain the soliton's shape after perturbations, and the agent generally cannot recover a disrupted pattern.

\section{Discussion}

We introduce a closed-loop framework for autonomous discovery and control of self-organizing phenomena, based on autotelic RL. Rather than setting initial conditions and passively observing outcomes, a goal-conditioned policy observes the evolving complex system and applies minimal, local perturbations toward diverse self-generated goals. We instantiate the framework on Lenia as a system named CARL, which discovers solitons across a wide range of update rules and procedurally generated initial states, generalizes to out-of-distribution conditions, and can efficiently map novel update rule spaces to identify regions that support solitons. Beyond discovery, CARL agents can also learn to control solitons by steering their movement direction. Finally, we demonstrate that CARL agents can serve as real-time interfaces, enabling human users to guide solitons through maze environments with simple directional commands.

A key design choice is the use of action costs that incentivize the agent to act sparsely, reflecting the principle that effective control of self-organizing systems should work alongside the system's intrinsic dynamics, not against them. In the mass tracking task, action costs lead the agent to discover self-sustaining solitons as a side effect of reward maximization — the cheapest way to maintain a target mass is to find a configuration that maintains itself. In the steering task, they produce a similar effect: instead of continuously micromanaging the soliton's trajectory, the agent learns to apply a brief perturbation that redirects it, then withdraws, allowing the soliton to continue along the new heading unassisted.

CARL 
generalizes well
to out-of-distribution conditions across variations in goals, action spaces, and novel update rules. This suggests the policies capture transferable system dynamics rather than overfitting. As a result, trained policies can be reused and composed to solve tasks beyond their original training objective. We demonstrate this through a maze-navigation task, where a human user modifies the agent’s goal (desired movement direction), while the policy handles the low-level control 
to achieve it in real time. This compositional reuse points toward 
functional integration, where distinct capabilities can be combined to solve increasingly complex tasks.
Such integration suggests a path toward hierarchical control, where higher-level agents or processes set subgoals for lower-level controllers.
We see CARL as a step toward artificial experimentalist frameworks, where agents not only learn how to autonomously act on complex systems, but also how to structure and combine those actions --- deciding \textit{what} to investigate through self-generated goals and \textit{how} to achieve it.

A key limitation is that instantiating the framework requires domain expertise: the reward function, action space, and observation design all encode knowledge about what makes a given system interesting. While the mass tracking objective sidestepped the need to define solitons explicitly, it still reflects a designer's intuition about Lenia. Domain expertise is inherent to scientific inquiry, but when the goal is to uncover phenomena we cannot yet characterize or anticipate, more open-ended approaches, such as intrinsic reward signals, adaptive goal sampling policies, or automated environment and task design, could reduce this dependence and broaden the scope of discovery.

While Lenia offers favorable conditions for closed-loop control --- full observability, determinism, simple action space --- extending CARL beyond such idealized systems is challenging: especially in biomedical and bioengineering settings, dynamics are partially observed, stochastic, high-dimensional, and multiscale in nature. 
Most biomedical efforts focus on micromanaging tangible targets --- single proteins, genes, or circuits --- but many biological systems are best understood not as static objects but as persistent, self-maintaining patterns across bioelectric, mechanical, metabolic, transcriptional, anatomical, and cognitive spaces, patterns that persist, grow, move, and reshape their surroundings \citep{mathewsCellularSignalingPathways2023, levin20254MultiscaleBody, fieldsThoughtsThinkersComplementarity2025a}. 
Thus, a central problem in biomedicine is to identify which low-level interventions can produce desired system-level outcomes, such as regeneration, cancer reprogramming, or rejuvenation \citep{Levin2021Cancer, PioLopez2025}.
In this sense, approaches like CARL point toward a new class of tools for treating self-organizing systems as programmable substrates, where high-level goals specified by the experimenter are interpreted by a layer of agents that infer multiscale interventions \citep{daviesSyntheticMorphologyAgential2023}.



\newpage
\section*{Acknowledgements}

We thank Barbora Hudcová for insightful discussions and guidance in navigating the Lenia Explorer dataset.
We thank members of the Flowers AI and CogSci Lab and the Levin Lab for helpful discussions.
We gratefully acknowledge support for this work provided through a sponsored research agreement with Astonishing Labs and from the Templeton World Charity Foundation, Inc. (Grant ID: TWCF-2021-20606). The opinions expressed in this publication are those of the authors and do not necessarily reflect the views of the funding agencies. The authors used generative AI tools to assist with writing the manuscript, developing experiment code and the companion website. All outputs were reviewed by the authors, who take full responsibility for the content.

\bibliographystyle{apalike}
\bibliography{references}


\end{document}